\documentclass[runningheads]{llncs}

\usepackage[table,dvipsnames]{xcolor}

\usepackage{graphicx}

\usepackage{booktabs}
\usepackage{multirow}
\usepackage{colortbl}

\usepackage{pifont}
\newcommand{\xmark}{\ding{55}} 

\usepackage{realboxes}     
\usepackage{algorithm}
\usepackage{algorithmic}
\usepackage{amsmath,amsfonts}
\usepackage{multicol}
\usepackage{subfig}        

\usepackage{graphicx}
\begin{document}
\title{Metric Matters: A Formal Evaluation of Similarity Measures in Active Learning for Cyber Threat Intelligence.}
\titlerunning{Similarity Measures in Active Learning for Cyber Threat Intelligence.}
%
\author{Sidahmed Benabderrahmane$^{*}$ \and Talal Rawhan }%
\authorrunning{S. Benabderrahmane et al.}
\institute{New York University, NYUAD, Division of Science. \\\email{sidahmed.benabderrahmane@gmail.com}}

\maketitle



 
\begin{abstract}
Advanced Persistent Threats (APTs) pose a severe challenge to cyber defense due to their stealthy behavior and the extreme class imbalance inherent in detection datasets. To address these issues, we propose a novel active learning-based anomaly detection framework that leverages similarity search to iteratively refine the decision space. Built upon an Attention-Based Autoencoder, our approach uses feature-space similarity to identify normal-like and anomaly-like instances, thereby enhancing model robustness with minimal oracle supervision. Crucially, we perform a formal evaluation of various similarity measures to understand their influence on sample selection and anomaly ranking effectiveness. Through experiments on diverse datasets, including DARPA Transparent Computing APT traces, we demonstrate that the choice of similarity metric significantly impacts model convergence, anomaly detection accuracy, and label efficiency. Our results offer actionable insights for selecting similarity functions in active learning pipelines tailored for threat intelligence and cyber defense.
\end{abstract}



\section{Introduction}

Anomaly detection plays a central role in cyber threat intelligence, particularly in identifying stealthy intrusions such as Advanced Persistent Threats (APTs) \cite{shenderovitz2024bon},\cite{singh2021novel}. These attacks are characterized by long-term, low-profile operations that mimic normal system behavior, making them especially challenging to detect \cite{xuan2024novel}. In real-world settings, datasets used for detecting such threats are highly imbalanced, where malicious behaviors are rare and heterogeneous, and the majority class overwhelmingly consists of routine system activity. Moreover, labeling such data often requires expert forensic analysis, making large-scale supervised learning approaches impractical due to the cost and latency of human-in-the-loop labeling.

Active learning has emerged as a promising paradigm to mitigate these challenges by selectively querying the most informative samples for labeling. This approach reduces the labeling burden while guiding the anomaly detection model to refine its decision boundaries over time. However, in the context of high-dimensional cybersecurity data, the effectiveness of active learning critically depends on the mechanism used to identify and prioritize these informative samples. In particular, \textit{similarity search within the feature space}---the process of identifying points that are structurally or semantically close to known examples---can significantly influence how the model explores the data and evolves its decision space.

Despite its importance, the impact of similarity measures in active learning for cyber threat detection remains underexplored. Traditional approaches often assume fixed similarity metrics or rely on simplistic heuristics that may not generalize across domains or data regimes. Yet, the choice of similarity metric inherently affects which samples are deemed ``similar'' to known anomalies or normals, and consequently, how the model updates its boundary between benign and malicious behavior.

In this paper, we address the following research question:\\
\textbf{How does the choice of similarity metric affect the performance of active learning-based anomaly detection in the context of cyber threat intelligence?}\\
To answer this, we build upon an Attention-Based Autoencoder AAE for anomaly detection and extend it with a modular active learning framework that incorporates various similarity measures to guide query selection. Specifically, our framework uses similarity search in the feature space to identify normal-like points for improving the model’s reconstruction capacity and anomaly-like points for refining the ranking of suspicious behaviors. These operations are iteratively integrated with oracle feedback to adaptively refine the decision space.

Our contributions are threefold:
\begin{itemize}
    \item \textbf{Framework Proposal:} We propose a novel active learning-based anomaly detection framework that leverages similarity search to iteratively refine the decision boundaries for APT detection.
    \item \textbf{Metric Evaluation:} We conduct a formal comparative analysis of six similarity measures---Hamming, Jaccard, Cosine, Dice, Euclidean, and a novel Normalized Matching 1s (NM1)---to assess their influence on active learning strategies.
    \item \textbf{Empirical Validation:} Through extensive experiments on several datasets---in cyber intelligence---we evaluate the impact of each similarity measure on detection scores, convergence speed, and label efficiency. Special focus is given to the DARPA Transparent Computing datasets, which include realistic APT scenarios.
\end{itemize}

The remainder of the paper is structured as follows. Section~2 reviews related work in anomaly detection, active learning, and similarity-based reasoning. Section~3 presents our proposed framework, including the similarity-guided strategies and similarity metric definitions. Section~4 details the experimental setup and results and discusses the observed trends and implications for cyber defense. Finally, Section~5 concludes with future directions.
\section{Related Work}
\subsection{Active Learning for Anomaly Detection}
Active learning is a learning paradigm that seeks to minimize labeling costs by strategically selecting the most informative unlabeled samples for annotation \cite{cacciarelli2024active,li2024survey}. In the context of anomaly detection, active learning is particularly appealing due to the scarcity of labeled anomalies and the high cost of expert feedback, especially in security-critical domains like APT detection \cite{chang2024multitask}.

Traditional active learning strategies include uncertainty sampling, where the model queries data points near its decision boundary, and diversity-based sampling, which aims to select a representative subset of the data to avoid redundancy. These strategies have shown promise in supervised and semi-supervised settings, and have been applied to domains such as intrusion detection, fraud detection, and medical diagnostics.

However, these strategies exhibit limitations when applied to high-dimensional, highly imbalanced datasets common in cybersecurity. First, model uncertainty becomes unreliable when the anomaly class is underrepresented, often leading to query selections that reinforce existing bias toward the normal class. Second, the curse of dimensionality can obscure meaningful structure in the data, reducing the effectiveness of purely uncertainty- or diversity-driven approaches. These limitations motivate the integration of structural information from the feature space—such as similarity relations between samples—to guide the active learning loop in a more informed and targeted manner.

\subsection{Similarity Measures in Machine Learning}

Similarity measures play a fundamental role in numerous machine learning tasks, including information retrieval, clustering, nearest-neighbor classification, and semi-supervised learning \cite{gupta2025comprehensive}. Metrics such as cosine similarity, Jaccard index, Hamming distance, and Euclidean distance serve as proxies for assessing closeness in feature space, which is often assumed to reflect semantic or behavioral similarity in the underlying domain \cite{grainger2025ai}.

In the context of anomaly detection, similarity is often used implicitly, for example in density-based methods like DBSCAN or isolation-based approaches where outliers are assumed to lie far from clusters of normal behavior. More recently, similarity functions have been integrated into deep learning pipelines for embedding learning, metric learning, and contrastive learning \cite{weller2014survey}.

Despite their wide applicability, the role of similarity measures in active learning for anomaly detection remains underexplored. Most existing frameworks adopt fixed or task-specific similarity functions without evaluating how the choice of metric influences the selection of query points or the refinement of decision boundaries. This gap is particularly salient in high-stakes domains like cybersecurity, where different similarity metrics can yield significantly different prioritization of suspicious behaviors. Understanding the behavior of these metrics under different data distributions is therefore critical for designing more effective and interpretable active learning strategies.

\subsection{ Cyber Threat Intelligence and APT Detection
}

Cyber threat intelligence (CTI) involves the acquisition, analysis, and dissemination of data related to malicious activity within or targeting information systems. One of the most challenging threat classes in this space is Advanced Persistent Threats (APTs)—stealthy, targeted attacks that operate over long durations while maintaining persistent access to compromised systems. Traditional intrusion detection systems often struggle to detect APTs due to their low-and-slow tactics and ability to mimic normal system behavior.

Modern anomaly detection methods have incorporated deep learning, graph-based modeling, and provenance analysis to detect subtle deviations associated with APTs. However, many of these methods rely on extensive labeled data or domain-specific heuristics that are not scalable in operational environments.

Within CTI workflows, the ability to prioritize threats based on similarity to known attacks is crucial. Analysts often rely on behavioral or structural similarity (e.g., process sequences, network flow patterns) to link new events to known Tactics, Techniques, and Procedures (TTPs). An active learning system that incorporates similarity-based prioritization can therefore serve as an effective decision-support tool, aligning model-driven discovery with analyst intuition. This motivation reinforces the need to study how different similarity measures shape the refinement of anomaly rankings and influence the active learning trajectory in the detection of APTs.
\section{Methodology}

\subsection{Overview of the Attention-Based Autoencoder Anomaly Detection Framework}

We propose an anomaly detection framework that integrates an attention-based autoencoder AAE with a similarity-guided active learning loop for detecting sophisticated cyber threats such as Advanced Persistent Threats (APTs). The model consists of two primary components: (i) a deep autoencoder architecture equipped with an attention mechanism for learning compact, interpretable representations of system behavior, and (ii) an active learning loop that iteratively queries the most informative samples based on their similarity to known labeled points. Let $x \in \mathbb{R}^d$ be a feature vector representing a system process or event trace. The encoder $E_\theta$ maps $x$ to a latent representation $z \in \mathbb{R}^k$ ($k<<d$), and the decoder $D_\phi$ reconstructs $\hat{x} = D_\phi(E_\theta(x))$. The reconstruction error is computed as: $\mathcal{L}_{\text{recon}}(x) = \|x - \hat{x}\|_2^2$. Anomalous points are assumed to yield higher reconstruction errors due to their deviation from the learned manifold of normal behavior. The attention mechanism $\alpha: \mathbb{R}^d \to [0,1]^d$ produces a soft weighting vector $\alpha(x)$, which is applied element-wise to the input: $x^{\ast} = x \odot \alpha(x)$, where $\odot$ denotes Hadamard (element-wise) product. This allows the model to focus on the most informative dimensions of the input during reconstruction.

The attention-based autoencoder AAE serves as the core anomaly detection model by learning a compact representation of normal system behavior and flagging deviations through reconstruction error. However, given the rarity and diversity of anomalies—particularly in APT scenarios—relying solely on unsupervised reconstruction is insufficient for high detection precision. To enhance the model’s discriminative power with minimal labeling effort, we embed the autoencoder within an active learning loop. In each iteration, reconstruction errors are used to rank unlabeled samples by their likelihood of being anomalous. A subset of top-ranked points is then queried to an oracle (human expert or a ground truth database), providing supervised feedback that guides the selection of similar points in the feature space. These similarity-guided updates help refine the decision space by both expanding the model's understanding of normal behavior and prioritizing anomaly-like regions for further exploration. This iterative integration of the autoencoder and similarity-aware querying enables more efficient and precise anomaly detection with a constrained labeling budget.
\subsection{Similarity-Based Active Learning Loop.}
The active learning component as explained in Algorithm 1 operates in iterative rounds. In each round, a ranked list of candidate anomalies is produced based on reconstruction errors. A subset of points is queried to an oracle (e.g., human analyst), which provides ground-truth labels indicating whether each queried point is normal or anomalous. These labels are then used to guide similarity search strategies within the feature space:

\begin{itemize}
    \item \textbf{Strategy 1 (Normal-Like Augmentation):} Points labeled as normal are used to find similar unlabeled points, which are assumed to be normal and added to the training set of AAE to improve reconstruction fidelity.

    \item \textbf{Strategy 2 (Anomalous-Like Prioritization):} Points labeled as anomalous are used to find similar points, which are prioritized in the next anomaly ranking.

    \item \textbf{Strategy 3 (Hybrid):} Combines both augmentation of normal-like points and prioritization of anomaly-like points to simultaneously improve model capacity and query targeting.
\end{itemize}

This iterative loop continues until a predefined stopping criterion is satisfied, which may include reaching a fixed number of queries, achieving convergence in the anomaly ranking list, or observing minimal changes in similarity scores across iterations. The rationale behind employing these adaptive querying and ranking strategies is to significantly reduce computational overhead and labeling effort, while still maintaining or improving the quality of anomaly prioritization by focusing only on the most informative samples.

\begin{algorithm}[t]
\caption{Active Learning with Similarity-Guided Attention Autoencoder}
\begin{algorithmic}[1]
\REQUIRE Dataset $\mathcal{X}$, Initial labeled subset $\mathcal{X}_l$, Unlabeled set $\mathcal{X}_u = \mathcal{X} \setminus \mathcal{X}_l$, Threshold $\tau$, Similarity metric $\text{sim}$, Oracle $\mathcal{O}$
\STATE Train attention autoencoder on $\mathcal{X}_l$
\FOR{each iteration $t = 1$ to $T$}
    \STATE Compute reconstruction errors $e(x)$ for all $x \in \mathcal{X}_u$
    \STATE Rank $\mathcal{X}_u$ in descending order of $e(x)$
    \STATE Query top-$k$ points $x_1, \dots, x_k$ to oracle $\mathcal{O}$
    \STATE Add labeled points to $\mathcal{X}_l$
    \IF{Strategy 1 applied}
        \STATE Identify normal-like points: $\text{sim}(x, n) \geq \rho$ for $n \in \mathcal{X}_l^{\text{normal}}$
        \STATE Add these to training set
    \ENDIF
    \IF{Strategy 2 applied}
        \STATE Identify anomaly-like points: $\text{sim}(x, a) \geq \xi$ for $a \in \mathcal{X}_l^{\text{anomaly}}$
        \STATE Prioritize these in ranking
    \ENDIF
    \IF{Hybrid strategy}
        \STATE Apply both normal-like and anomaly-like updates
    \ENDIF
    \STATE Retrain attention autoencoder on updated $\mathcal{X}_l$
\ENDFOR
\RETURN Final anomaly ranking
\end{algorithmic}
\end{algorithm}
\section{Data Representation and Similarity Measures}

\subsection{Binary Data Structure}

We begin by describing the structure of the data used in our anomaly detection framework, followed by the similarity metrics applied to quantify relationships between data points. These similarity measures are foundational to our ranking mechanism and play a central role in prioritizing candidate anomalies.

We consider a binary dataset \( X \in \{0,1\}^{n \times m} \), where:
\begin{itemize}
    \item \( n \) is the number of data points (rows),
    \item \( m \) is the number of binary features (columns).
\end{itemize}

Each entry \( x_{i,j} \in \{0, 1\} \) indicates whether the \( j \)-th feature is active (\(1\)) or inactive (\(0\)) for the \( i \)-th data point. For instance, in a cybersecurity context, each row \( i \) may represent a system process, while each column \( j \) corresponds to an event such as file access, network activity, or command execution.

Consider the example below, where \( n = 3 \) and \( m = 5 \):

\[
X = \begin{bmatrix}  
1 & 0 & 1 & 1 & 0 \\
0 & 1 & 1 & 1 & 0 \\  
1 & 1 & 1 & 0 & 0
\end{bmatrix}
\]

This dataset represents three samples:
\begin{itemize}
    \item Data point 1 has features 1, 3, and 4 active.
    \item Data point 2 has features 2, 3, and 4 active.
    \item Data point 3 has features 1, 2, and 3 active.
\end{itemize}

Our goal is to compute similarities between pairs of rows in \( X \) to identify structural patterns and surface anomalous behaviors.

\subsection{Evaluated Similarity Measures}

We evaluate six widely used similarity metrics, each with distinct assumptions and behavior over binary data. These are summarized in Table~\ref{simeval} and formally defined as follows:

\begin{itemize}
    \item \textbf{Hamming Similarity}: Measures agreement across all feature positions.
    \[
    \text{sim}_{\text{Hamming}}(x, x') = 1 - \frac{1}{m} \sum_{i=1}^m \mathbb{I}[x_i \neq x'_i]
    \]

    \item \textbf{Jaccard Index}: Measures the intersection over union of active features.
    \[
    \text{sim}_{\text{Jaccard}}(x, x') = \frac{|x \land x'|}{|x \lor x'|}
    \]

    \item \textbf{Cosine Similarity}: Measures the cosine of the angle between vectors.
    \[
    \text{sim}_{\cos}(x, x') = \frac{x \cdot x'}{\|x\| \cdot \|x'\|}
    \]

    \item \textbf{Dice Coefficient}: A symmetric alternative to Jaccard emphasizing matching 1s.
    \[
    \text{sim}_{\text{Dice}}(x, x') = \frac{2|x \land x'|}{|x| + |x'|}
    \]

    \item \textbf{Euclidean Similarity}: A Gaussian kernel applied to the Euclidean distance.
    \[
    \text{sim}_{\text{Euclid}}(x, x') = \exp\left(-\frac{\|x - x'\|_2^2}{\sigma^2}\right)
    \]

    \item \textbf{Normalized Matching 1s (NM1)}: We introduce here NM1, that focuses exclusively on shared active features (1s), excluding common inactives. It is especially suited for sparse, binary datasets where active events carry more semantic weight.

    For two binary vectors \( A \) and \( B \), we define:
    \[
    \text{sim}_{\text{NM1}}(x, x') = \frac{|x \land x'|}{\max(|x|, |x'|)}
    \]
    where:
    \begin{itemize}
        \item \( |A  \land B| \) is the number of indices where both vectors have 1s,
        \item \( |A| \), \( |B| \) are the total number of 1s in \( A \) and \( B \), respectively.
    \end{itemize}
\end{itemize}

Each metric captures a different notion of similarity. For instance, Hamming considers full bitwise alignment, while NM1 emphasizes co-activated features—particularly important in anomaly detection contexts where the presence (rather than absence) of rare actions is key.

\begin{table}[H]
\centering
\tiny
\caption{Theoretical Properties of Evaluated Similarity Measures}
\label{simeval}
\begin{tabular}{lcccc}
\toprule
\textbf{Metric} & \textbf{Symmetric} & \textbf{Binary-Only} & \textbf{Sparsity Robust} & \textbf{Cost} \\
\midrule
Hamming & \checkmark & \checkmark & Medium & Low \\
Jaccard & \checkmark & \checkmark & High & Low \\
Cosine & \checkmark & \xmark & High & Medium \\
Dice & \checkmark & \checkmark & High & Low \\
Euclidean & \checkmark & \xmark & Low & High \\
NM1 & \checkmark & \checkmark & Very High & Low \\
\bottomrule
\end{tabular}
\end{table}

\subsection{Experimental Setup}
\subsubsection{Dataset Description}
We use real-world APT datasets with binary process behavior vectors. Each vector represents the presence/absence of key system actions. APTs are rare and labeled via known attack stages (recon, exploit, persist, etc.). These cyber security data sources used in this paper come from the Defense Advanced Research Projects Agency (DARPA)’s \verb|Transparent| \verb|Computing TC|\footnote{https://gitlab.com/adaptdata} program~\cite{darpa}. The aim of this program is to provide transparent provenance data of system activities and component interactions across different operating systems (OS) and spanning all layers of software abstractions. Specifically, the datasets include system-level data, background activities, and system operations recorded while APT-style attacks are being carried out on the underlying systems. Preserving the provenance of all system elements allows for tracking the interactions and dependencies among components. Such an interdependent view of system operations is helpful for detecting activities that are individually legitimate or benign but collectively might indicate abnormal behavior or malicious intent. Here we specifically employ DARPA’s data that has undergone processing conducted by the \verb|ADAPT| (Automatic Detection of Advanced Persistent Threats) project’s ingester. The records come from four different source OS, namely Android (called in the TC program Clearscope), Linux (called Trace), BSD (called Cadets), and Windows (called Fivedirections or 5dir). For each system, the data comes from two separate attack scenarios: scenario E1 (also called Pandex) and scenario E2 (called Bovia), respectively. The processing includes ingesting provenance graph data into a graph database as well as additional data integration and deduplication steps. The final data includes a number of Boolean-valued datasets (data aspects)
, with each representing an aspect of the behavior of system processes. Each row in such a data aspect 
is a data point representing a single process run on the respective OS. It is expressed as a Boolean vector whereby a value of 1 in a vector cell indicates the corresponding attribute applies to that process. 

Specifically, the relevant 
datasets are interpreted as follows:

\begin{itemize}
    \item \verb|ProcessEvent| (PE): Its attributes are event types performed by the processes. A value of 1 in \verb|process[i]| means the process has performed at least one event of type $i$.
    \item \verb|ProcessExec| (PX): The attributes are executable names that are used to start the processes.
    \item \verb|ProcessParent| (PP): Its attributes are executable names that are used to start the parents of the processes.
    \item \verb|ProcessNetflow| (PN): The attributes here represent IP addresses and port names that have been accessed by the processes.
    \item \verb|ProcessAll| (PA): This dataset is described by the disjoint union of all attribute sets from the previous datasets.
\end{itemize}
Overall, with two attack scenarios, four OS (BSD, Windows, Linux, Android) and five aspects (PE, PX, PP, PN, PA), a total of forty individual datasets are composed, as illustrated in Figure 1. 
 \begin{figure}[h]
 
    \centering
    \includegraphics[width=0.2\linewidth]{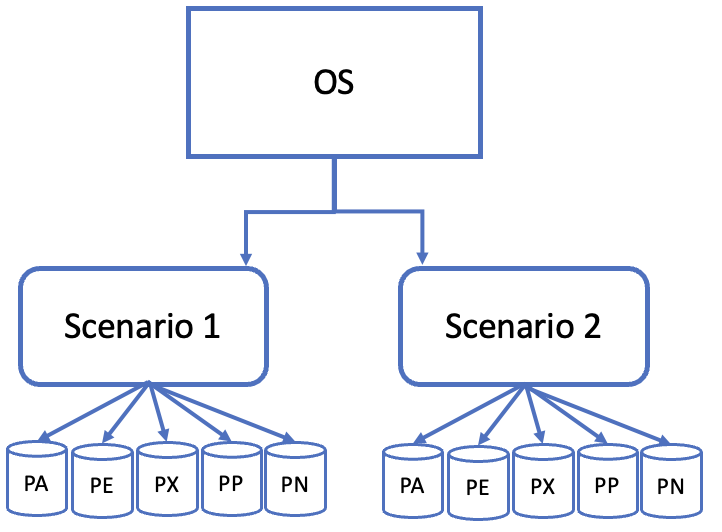}
    \caption{Organization of the DARPA's TC datasets. Each OS undergoes two attack scenarios, each of which contains five data aspects ets. With four OS (BSD, Windows, Linux, Android), two attack scenarios, and five aspects (PE, PX, PP, PN, PA), a total of forty individual datasets are composed. Each forensic configuration (OS$\times$attack scenario$\times$data aspect) represents a single dataset.}
    \label{fig:DARPATC}

 \end{figure}
 
They are described in Table~\ref{datatable} whereby the last column provides the number of attacks in each dataset. The substantially imbalanced nature of the datasets is clearly seen here. For each forensic configuration (OS$\times$attack scenario$\times$data aspect) we have the number of processes (instances) and the corresponding events (features). For instance, $Windows\_E1\_PE$ is the dataset represented by $PE$ aspect belonging to Windows OS, produced during the first attack scenario E1 (Pandex). It contains 17569 instances and 22 features with a total of 8 APTs anomalies (0.04\%). 

\begin{table}
\caption{Summary of the first source of 40 benchmark datasets belonging to DARPA's TC program for APT detection. A dataset entry (columns 4 to 8) is described by a number of rows (processes) / number of columns (attributes). For instance, with ProcessAll (PA) obtained from the second scenario using Linux, the dataset has 282104 rows and 6435 attributes with 46 APT attacks (0.01\%). }
\centering
\resizebox{0.99\textwidth}{!}{
\begin{tabular}{|l|l||l|l|l|l|l|l|l|l|}
\hline & Scenario & Size& $PE$   & $PX$  & $PP$  & $PN$     & $PA$  & $nb\_attacks$    & $\%\frac{nb\_attacks}{nb\_processes}$     \\ \hline \hline
BSD    & 1 &288 MB &76903 / 29  & 76698 / 107  & 76455 / 24  & 31 / 136  & 76903 / 296 & 13&0.02\\  
    & 2 &1.27 GB &224624 / 31  &224246 / 135  & 223780 / 37  & 42888 / 62 &  224624 / 265      & 11&0.004\\ \hline
Windows & 1 &743 MB & 17569 / 22    &  17552 / 215  &   14007 / 77        &   92 / 13963      & 17569 / 14431& 8&0.04\\  
   & 2 &9.53 GB& 11151 / 30    &  11077 / 388  & 10922 / 84  & 329 / 125      &  11151 / 606    &8&0.07\\ \hline
Linux  & 1 &2858 MB &247160 / 24 & 186726 / 154 & 173211 / 40 & 3125 / 81 & 247160 / 299  &25&0.01\\
    & 2 &25.9 GB &282087 / 25 & 271088 / 140 & 263730 / 45 &6589 / 6225 &  282104 / 6435      &46&0.01\\ \hline
Android& 1 &2688 MB&102 / 21     &102 / 42&0 / 0&8 / 17& 102 / 80&9&8.8\\
&2 &10.9 GB&12106 / 27     &12106 / 44&0 / 0&4550 / 213&12106 / 295 &13&0.10\\ \hline
\end{tabular}
}

 \label{datatable}
\end{table}

\subsection{Metrics:} To evaluate the quality of anomaly ranking produced during active learning, we adopt the Normalized Discounted Cumulative Gain (nDCG) as our primary performance metric. Unlike traditional metrics such as the Area Under the Receiver Operating Characteristic Curve (AUC), which measures classification performance over all thresholds, nDCG explicitly evaluates the ranking order of predicted anomalies—placing greater emphasis on correctly identifying true anomalies at the top of the ranked list. This is particularly advantageous in highly imbalanced settings, such as APT detection, where the number of anomalies is extremely small relative to normal instances. AUC can often overestimate model performance by assigning equal importance to true positives and false positives across all decision thresholds, thereby failing to capture the practical need to surface the most critical anomalies early. In contrast, nDCG reflects the operational priorities of security analysts by rewarding models that successfully concentrate true anomalies in the top-k results. This makes nDCG a more suitable and informative metric for evaluating the effectiveness of active learning strategies in cyber threat intelligence contexts.

The Normalized Discounted Cumulative Gain (nDCG), defined as:$\text{nDCG}_k = \frac{\text{DCG}_k}{\text{IDCG}_k}$, where the DCG at rank $k$ is given by: $\text{DCG}_k = \sum_{i=1}^k \frac{2^{r_i} - 1}{\log_2(i + 1)}$ and $r_i \in \{0, 1\}$ is the relevance label at rank position $i$ (1 if the $i$-th item is an anomaly, 0 otherwise). The ideal DCG (IDCG) is the DCG of the optimal ranking (i.e., all anomalies ranked at the top) and serves as a normalization factor: $\text{IDCG}_k = \sum_{i=1}^{k'} \frac{2^1 - 1}{\log_2(i + 1)}$, where $k'$ is the number of actual anomalies in the top-$k$ ground truth ranking. The resulting nDCG value ranges from 0 to 1, with 1 indicating a perfectly ranked list.

\subsection{Protocol:} We run a fixed-budget active learning loop over 20 iterations. In each iteration:
\begin{enumerate}
    \item Anomaly scores are computed using reconstruction errors.
    \item Top-ranked samples are queried to the oracle.
    \item Similarity search is performed using the selected similarity metric.
    \item Augmented training sets are constructed using similarity-based strategies.
    \item The model is retrained with the updated labeled dataset.
\end{enumerate}

\section{Experimental Results}

This section presents an empirical evaluation of the proposed attention-based autoencoder framework combined with similarity-guided active learning. Our focus is to assess how the choice of similarity metric influences anomaly ranking performance, query efficiency, and model refinement. All experiments were conducted under a fixed oracle budget of 20 iterations, and results were averaged over multiple random seeds to ensure robustness.

\subsection{Overall Ranking Performance}

We first assess the overall effectiveness of each similarity metric in improving anomaly ranking, as measured by nDCG across all datasets and strategies. Results show clear variation in performance depending on the similarity function used.

\begin{figure}[h]
    \centering
    \includegraphics[width=0.45\linewidth]{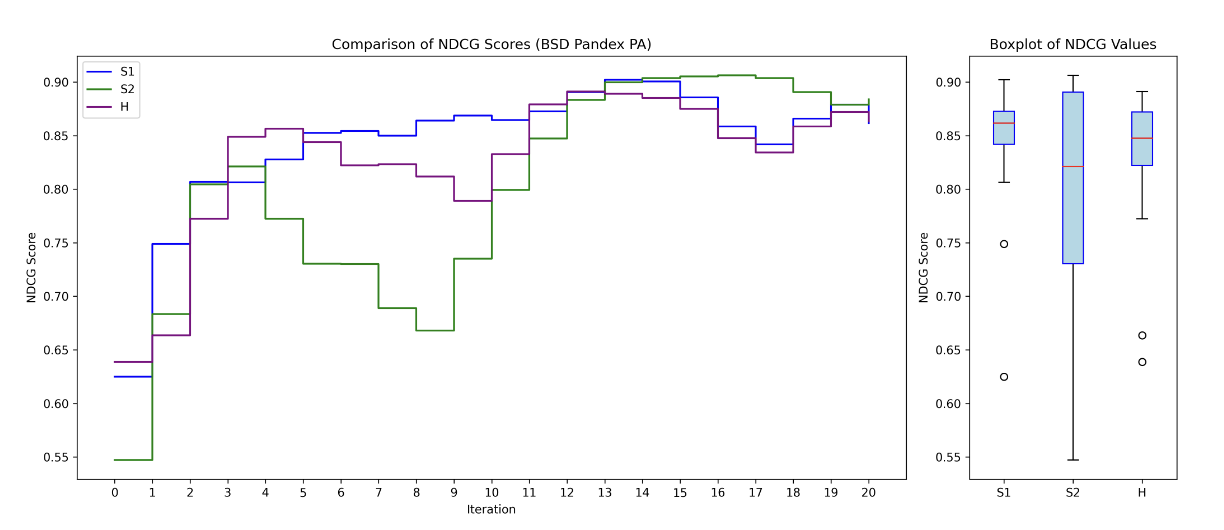}
     \includegraphics[width=0.45\linewidth]{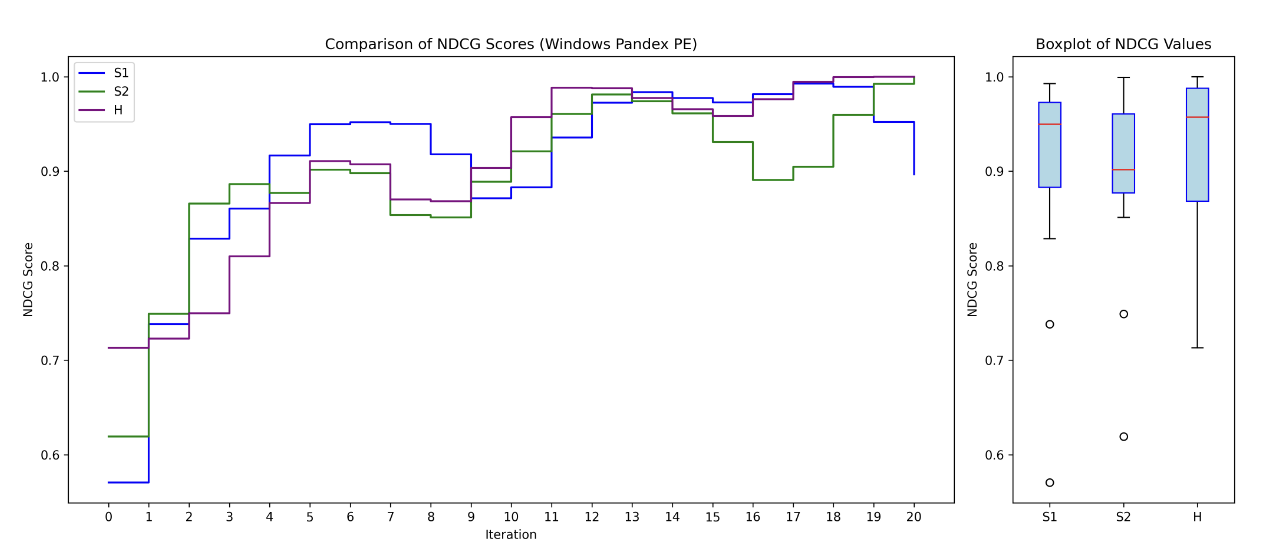}
    \caption{%
Comparison of nDCG scores over 20 active learning iterations using the Normalized Matching 1s (NM1) similarity metric under three query selection strategies: S1 (normal-like similarity), S2 (anomaly-like similarity), and H (hybrid). The left subplots show the progression of nDCG scores during learning, while the right subplots summarize the score distributions using boxplots. The results highlight differences in convergence behavior and ranking stability across strategies. Left: BSD PA dataset, Right: Windows PE dataset.%
}
    \label{fig:ndcgvariation}
\end{figure}

Figure~\ref{fig:ndcgvariation} illustrates the evolution of nDCG scores across 20 active learning iterations using the proposed Normalized Matching 1s (NM1) similarity metric under three different query selection strategies: Strategy 1 (normal-like similarity), Strategy 2 (anomaly-like similarity), and Strategy 3 (hybrid combination). Due to space limitations, we only present active learning figures using NM1. The line plot on the left reveals the progressive improvement in ranking quality as labeled feedback is incorporated, while the boxplot on the right summarizes the distribution of nDCG scores across all iterations for each strategy. We observe that Strategy 1 exhibits steady and consistent improvement, whereas Strategy 2, while achieving higher peak performance, shows greater variability across iterations. The hybrid strategy combines the strengths of both approaches, yielding stable and high-ranking results with reduced fluctuation.

To generalize this insight, we will extend this analysis by generating equivalent nDCG trajectories and summary statistics for all other evaluated similarity measures and datasets. For each metric, we will compute and compare the average nDCG across iterations for the three strategies (from the boxplots), providing a comprehensive understanding of how similarity choices interact with active learning dynamics. This comparative study will offer empirical guidance for selecting the most effective similarity-strategy combinations in real-world anomaly detection workflows.
\vspace{-1 em}
\subsection{Heatmap Analysis: nDCG Scores Across Datasets, Similarity Measures, and Strategies}

\begin{figure*}[h]
    \centering
    \includegraphics[width=1\linewidth]{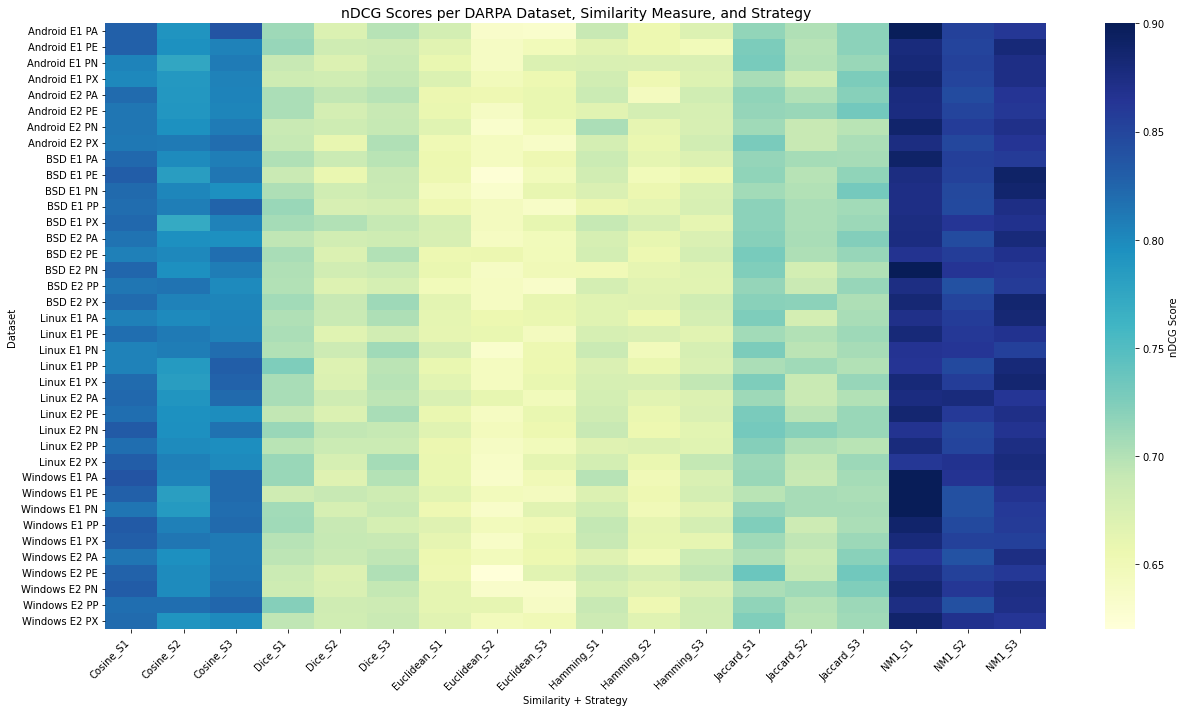}
    \caption{%
Heatmap of nDCG scores across DARPA dataset configurations (5 OS $\times$ 2 scenarios $\times$ 5 feature sets) under 18 combinations of similarity measures and active learning strategies. Each cell represents the average nDCG score achieved by a given similarity-strategy pairing on a specific dataset. Darker shades indicate better ranking performance. NM1 consistently yields the highest scores, followed by Cosine, while traditional metrics like Euclidean and Hamming perform poorly in sparse, high-dimensional anomaly detection tasks.%
}

    \label{fig:SimilaritySearchHeatmap}
\end{figure*}

Figure~\ref{fig:SimilaritySearchHeatmap} presents a heatmap visualization of nDCG scores obtained across the DARPA dataset configurations under 18 combinations of similarity measures and active learning strategies (6 similarities, 3 strategies). Each cell represents the average nDCG score resulting from a specific similarity-strategy pairing applied to a given dataset (c.f. average values in boxplots). Darker shades (blue) indicate higher nDCG values, reflecting better anomaly ranking performance.

\paragraph{Interpretation.}
The rightmost columns, corresponding to the Normalized Matching 1s (NM1) similarity under Strategies S1, S2, and S3, consistently display the darkest shades across almost all datasets. This confirms that NM1 offers the strongest and most stable performance for anomaly ranking in the context of sparse, binary cybersecurity data.

Cosine similarity, shown in the leftmost columns, also achieves competitive performance—especially under Strategy 1 and Strategy 3. These configurations exhibit moderately high nDCG scores across several datasets, though with slightly greater variance than NM1.

In contrast, similarity measures such as Jaccard, Dice, Hamming, and Euclidean yield noticeably lighter shades in the heatmap. Their performance is less consistent and generally lower across the board, particularly for Strategy 2, suggesting a lack of robustness in guiding effective query selection.

\paragraph{Strategic Patterns.}
When comparing the three strategies:
\begin{itemize}
    \item \textbf{Strategy 1 (normal-like augmentation)} consistently produces strong and stable rankings, particularly when paired with NM1 or Cosine.
    \item \textbf{Strategy 2 (anomaly-like prioritization)} shows higher performance variability, indicating sensitivity to the underlying similarity function and data distribution.
    \item \textbf{Strategy 3 (hybrid)} offers a balanced compromise, often enhancing stability while preserving high ranking performance, especially for NM1.
\end{itemize}

\paragraph{Key Takeaways.}
\begin{itemize}
    \item \textbf{NM1 is the most effective similarity measure} for active learning-based anomaly detection, achieving superior nDCG scores across nearly all datasets and strategies.
    \item \textbf{Cosine similarity ranks second}, offering good generalization and high early-stage ranking accuracy, particularly with Strategy 1.
    \item \textbf{Traditional similarity measures} such as Jaccard, Dice, Hamming, and Euclidean are less suited to high-dimensional binary cybersecurity data and fail to deliver competitive rankings.
    \item \textbf{Strategy 1} appears to provide the most consistent performance, especially when paired with similarity functions that exploit normal structure in the data.
    \item \textbf{Strategy 2} is more sensitive to noise and label sparsity, requiring more robust similarity definitions to avoid degraded performance.
\end{itemize}

These findings empirically validate the importance of selecting appropriate similarity measures and active learning strategies in the design of cyber threat detection pipelines. The consistent performance of NM1 further reinforces its domain suitability for detecting APTs in sparse binary environments.

\vspace{-3 em}
 \section{Conclusion}

In this work, we investigated the impact of similarity measures on active learning-based anomaly detection for cyber threat intelligence, with a particular focus on detecting Advanced Persistent Threats (APTs) in high-dimensional, sparse, and imbalanced datasets. We proposed a modular framework that integrates an attention-based autoencoder with a similarity-guided active learning loop to iteratively refine the anomaly ranking process. Our framework explored three querying strategies—normal-like augmentation, anomaly-like prioritization, and a hybrid combination—paired with six widely used similarity measures. We conducted a comprehensive evaluation across several DARPA Transparent Computing datasets, capturing multiple operating systems, attack scenarios, and binary feature abstractions. The evaluation used nDCG as a ranking-aware metric to assess the quality and efficiency of each similarity-strategy combination. The results demonstrate that the choice of similarity measure significantly affects the performance of active learning. The Normalized Matching 1s (NM1) metric consistently outperformed all other similarity functions, achieving high and stable nDCG scores across all strategies and datasets. Cosine similarity emerged as a viable second-best option, especially under Strategy 1. In contrast, traditional measures such as Hamming, Euclidean, and Dice showed limited effectiveness in sparse binary settings. Our findings underscore the importance of metric selection in designing robust and efficient anomaly detection pipelines for cyber defense. They also provide practical guidance for aligning similarity functions with querying strategies to optimize threat prioritization and label efficiency. In future research, we plan to extend this study by incorporating learnable similarity functions via contrastive or metric learning, evaluating performance in streaming and federated learning settings, and integrating semantic similarity using threat ontologies to enhance interpretability and analyst trust in decision support systems.
\section*{Disclosure of Interests:} The authors declare that they have no conflict of interest.
 
\bibliographystyle{IEEEtran}
\bibliography{output}

\begin{thebibliography}{10}
\providecommand{\url}[1]{#1}
\csname url@samestyle\endcsname
\providecommand{\newblock}{\relax}
\providecommand{\bibinfo}[2]{#2}
\providecommand{\BIBentrySTDinterwordspacing}{\spaceskip=0pt\relax}
\providecommand{\BIBentryALTinterwordstretchfactor}{4}
\providecommand{\BIBentryALTinterwordspacing}{\spaceskip=\fontdimen2\font plus
\BIBentryALTinterwordstretchfactor\fontdimen3\font minus \fontdimen4\font\relax}
\providecommand{\BIBforeignlanguage}[2]{{%
\expandafter\ifx\csname l@#1\endcsname\relax
\typeout{** WARNING: IEEEtran.bst: No hyphenation pattern has been}%
\typeout{** loaded for the language `#1'. Using the pattern for}%
\typeout{** the default language instead.}%
\else
\language=\csname l@#1\endcsname
\fi
#2}}
\providecommand{\BIBdecl}{\relax}
\BIBdecl

\bibitem{shenderovitz2024bon}
G.~Shenderovitz and N.~Nissim, ``Bon apt: Detection, attribution, and explainability of apt malware using temporal segmentation of api calls,'' \emph{Computers \& Security}, vol. 142, p. 103862, 2024.

\bibitem{singh2021novel}
R.~Singh and G.~Srivastav, ``Novel framework for anomaly detection using machine learning technique on cic-ids2017 dataset,'' in \emph{2021 International Conference on Technological Advancements and Innovations (ICTAI)}.\hskip 1em plus 0.5em minus 0.4em\relax IEEE, 2021, pp. 632--636.

\bibitem{xuan2024novel}
C.~D. Xuan and T.~T. Nguyen, ``A novel approach for apt attack detection based on an advanced computing,'' \emph{Scientific Reports}, vol.~14, no.~1, p. 22223, 2024.

\bibitem{cacciarelli2024active}
D.~Cacciarelli and M.~Kulahci, ``Active learning for data streams: a survey,'' \emph{Machine Learning}, vol. 113, no.~1, pp. 185--239, 2024.

\bibitem{li2024survey}
D.~Li, Z.~Wang, Y.~Chen, R.~Jiang, W.~Ding, and M.~Okumura, ``A survey on deep active learning: Recent advances and new frontiers,'' \emph{IEEE Transactions on Neural Networks and Learning Systems}, 2024.

\bibitem{chang2024multitask}
W.~Chang, K.~Liu, K.~Ding, P.~S. Yu, and J.~Yu, ``Multitask active learning for graph anomaly detection,'' \emph{arXiv preprint arXiv:2401.13210}, 2024.

\bibitem{gupta2025comprehensive}
S.~Gupta, U.~Thakar, and S.~Tokekar, ``A comprehensive survey on techniques for numerical similarity measurement,'' \emph{Expert Systems with Applications}, p. 127235, 2025.

\bibitem{grainger2025ai}
T.~Grainger, D.~Turnbull, and M.~Irwin, \emph{AI-Powered Search}.\hskip 1em plus 0.5em minus 0.4em\relax Simon and Schuster, 2025.

\bibitem{weller2014survey}
D.~J. Weller-Fahy, B.~J. Borghetti, and A.~A. Sodemann, ``A survey of distance and similarity measures used within network intrusion anomaly detection,'' \emph{IEEE Communications Surveys \& Tutorials}, vol.~17, no.~1, pp. 70--91, 2014.

\bibitem{darpa}
``Tran. computing,'' https://www.darpa.mil/program/transparent-computing.

\end{thebibliography}

\end{document}